\RequirePackage{fix-cm}
\documentclass{svjour3_arxiv}                     % onecolumn (standard format)
\smartqed  % flush right qed marks, e.g. at end of proof
\usepackage{graphicx}

\usepackage{natbib}
\usepackage{macros}

\graphicspath{{figures/}}
\begin{document}

\title{Statistical applications of contrastive learning%\thanks{Grants or other notes
}
\subtitle{}


%\author{}
\author{Michael U. Gutmann \and Steven Kleinegesse \and Benjamin Rhodes}


\institute{M.U.\ Gutmann \at
  School of Informatics\\
  University of Edinburgh\\
%  Tel.: +44 (0) 131 650 5190\\
  \email{michael.gutmann@ed.ac.uk}           %  \\
  \and
  S.\ Kleinegesse \at
  School of Informatics\\
  University of Edinburgh\\
  \email{steven.kleinegesse@ed.ac.uk}
  \and
  B.\ Rhodes
  \at
  School of Informatics\\
  University of Edinburgh\\
  \email{ben.rhodes@ed.ac.uk}
}

\date{Received: date / Accepted: date}

\maketitle

\begin{abstract}
The likelihood function plays a crucial role in statistical inference
and experimental design. However, it is computationally intractable
for several important classes of statistical models, including
energy-based models and simulator-based models. Contrastive learning
is an intuitive and computationally feasible alternative to
likelihood-based learning. We here first provide an introduction to
contrastive learning and then show how we can use it to derive methods
for diverse statistical problems, namely parameter
estimation for energy-based models, Bayesian inference for
simulator-based models, as well as experimental design.
\keywords{Contrastive learning \and energy-based models \and simulator-based models \and parameter estimation \and Bayesian inference \and Bayesian experimental design}
%\subclass{MSC code1 \and MSC code2 \and more}
\end{abstract}

\section{Introduction}
\label{sec:intro}
Contrastive or self-supervised learning is an intuitive learning
principle that is being used with much success in a broad range of
domains, e.g.\ natural language processing \citep{Mnih2013b,
  Kong2020a}, image modelling \citep{Gutmann2013, Aneja2021a} and
representation learning \citep{Gutmann2009b, vandenOord2018, Chen2020} to name a
few. It is a computationally feasible yet statistically principled
alternative to likelihood-based learning when the likelihood function
is too expensive to compute and thus has wide applicability.

In this paper we focus on the statistical side of contrastive learning
rather than on a particular application domain. We first explain the
principles of contrastive learning and then show how we can use it to
solve a diverse set of difficult statistical tasks, namely (1)
parameter estimation for energy-based models, (2) Bayesian inference
for simulator-based models, as well as (3) experimental design. We will
introduce these problems in detail and explain when and why
likelihood-based learning becomes computationally infeasible. The
three problems involve different models as well as tasks---inference
versus experimental design. They are thus notably different from each
other, which highlights the broad usage of contrastive learning. Yet,
they share common technical barriers which we will work out and show
how contrastive learning tackles them.

We focus on contrastive learning, but this should not give the
impression that it is the only statistical technique that may be used
to deal with any one of the three problems mentioned above. We will not
have space to review alternatives in detail and refer the reader in the relevant
sections to related work that deals with each of the statistical
problems on their own.

The paper is organised as follows: In Section \ref{sec:background}, we
provide background on likelihood-based learning and experimental
design, and then introduce energy-based and simulator-based models,
explaining why they both typically lead to intractable likelihood
functions. In Section \ref{sec:contrastive-learning}, we introduce
contrastive learning, explaining the basic idea and introducing its
two main ingredients, namely the loss function and the construction of
the contrasting reference data. In Section \ref{sec:applications}, we
then apply contrastive learning to the three aforementioned statistical
problems.

\section{Computational issues with likelihood-based learning and design}
\label{sec:background}
We first briefly review the likelihood function and its use in
learning and experimental design. We then introduce two different
classes of statistical models, namely energy-based models and
simulator-based models, and explain why their likelihood function is
typically intractable.

\subsection{Likelihood function}
The likelihood function $L(\thetab)$ is classically the main workhorse to 
solve inference and design tasks. Loosely speaking, it is
proportional to the probability that the model generates data $\x$ that is similar to the observed data $\xobs$ when using parameter value $\thetab$. Here, $\x$, as well as as $\xobs$, denotes generic data that may be a collection of independent data points or e.g.\ a time series. More formally, the likelihood function can be
expressed as the limit
\begin{equation}
  L(\thetab) = \lim_{\epsilon \to 0} c_{\epsilon} \Pr(\x \in B_{\epsilon}(\xobs) | \thetab),
\end{equation}
where $B_{\epsilon}(\xobs)$ is an $\epsilon$-ball around the observed data
$\xobs$ and $c_\epsilon$ is a normalising term that ensures that
$L(\thetab)$ is non-zero if $\Pr(\x \in B_{\epsilon}(\xobs) |
\thetab)$ tends to zero for $\epsilon \to 0$. 

For models expressed as a family of probability density functions
(pdfs) $\{p(\x|\thetab)\}$ indexed by $\thetab$, the likelihood
function is simply given by
\begin{equation}
  L(\thetab) \propto p(\xobs | \thetab).
\end{equation}
For maximum likelihood estimation, we then solve the optimisation
problem
\begin{align}
  \hat{\thetab} &= \argmax_{\thetab} p(\xobs|\thetab)
\end{align}
while for Bayesian inference with a prior pdf $p(\thetab)$, we compute the conditional pdf
\begin{align}
p(\thetab|\xobs) &= \frac{p(\xobs |\thetab)}{p(\xobs)} p(\thetab), & p(\xobs) &= \int p(\xobs|\thetab) p(\thetab) d \thetab,
\end{align}
or sample from it via Markov chain Monte Carlo \citep[MCMC, e.g.][]{Green2015}. Both maximum likelihood estimation and Bayesian
inference are generally computationally intensive unless
$p(\x|\thetab)$ exhibits structure (e.g.\ independencies or functional
forms) that can be exploited.

For experimental design, we include a dependency on
the design variable $\d$ in our model, so that the family of model pdfs is
given by $\{p(\x|\thetab, \d)\}$. Whilst there are multiple approaches to experimental design \citep{Chaloner1995, Ryan2016}, we here consider the case of Bayesian experimental design for parameter estimation with the mutual information between data and parameters as the utility function. The design task is then to determine the values of $\d$ that achieve maximal mutual information, i.e.\ to determine the design  $\d$ that is likely to yield data that is most informative about the parameters. The design problem can be expressed as the following optimisation problem
\begin{align}
  \hat{\d} &= \argmax_{\d} \MI_{\d}(\x, \thetab),\\ \MI_{\d}(\x,
  \thetab)&= \KL\left(p(\thetab, \x | \d) || p(\thetab | \d) p(\x|\d)\right)  \\ &= \E_{p(\x, \thetab|\d)} \log \left[
    \frac{p(\x |\thetab, \d)}{p(\x|\d)}\right],
  \label{eq:argmaxMI}
\end{align}
where $\KL$ denotes the Kullback-Leibler divergence and $\MI_\d$ the
mutual information for design $\d$. The mutual information is defined
via expectations (integrals), namely the outer expectation over the
joint $p(\x, \thetab|\d)$ as well as the expectation defining the
marginal $p(\x|\d)$ in the denominator, i.e.\
\begin{align}
  p(\x|\d) = \int p(\x | \thetab, \d) p(\thetab) d\thetab,
\end{align}
where we have assumed that the prior $p(\thetab)$ does not depend on
the design $\d$ as commonly done in experimental design. These expectations usually need to be approximated, e.g.\ via a
sample average.

Learning, i.e.\ parameter estimation or Bayesian inference, and
experimental design typically have high computational cost for any
given family of pdfs $p(\x | \thetab)$ or $p(\x | \thetab,
\d)$. However, not all statistical models are specified in terms of a
family of pdfs. Two important classes that we deal with in this paper
are the energy-based and the simulator-based models (also called
unnormalised and implicit models, respectively). The two models are
rather different but their common point is that high-dimensional
integrals needed to be computed to evaluate their pdf, and hence the
likelihood. These integrals render standard likelihood-based learning
and experimental design as reviewed above computationally intractable.

\subsection{Energy-based model}
\label{sec:EBM}
Energy-based models (EBMs) are used in various domains, for instance
to model images \citep[e.g.][]{Gutmann2013, Du2019, Song2019} or
natural language \citep[e.g.][]{Mnih2013b, Mikolov2013}, and have
applications beyond modelling in out-of-distribution detection
\citep{Liu2020}. They are specified by a real-valued energy function
$E(\x; \thetab)$ that defines the model up to a proportionality factor, i.e.\
\begin{align}
  p(\x| \thetab) &\propto \phi(\x; \thetab), &  \phi(\x; \thetab)&= \exp(-E(\x; \thetab)).
\end{align}
The exponential transform ensures that $\phi(\x; \thetab)\ge 0$ and strict inequality holds for finite energies. Note that larger values
of $E(\x; \thetab)$ correspond to smaller values of $p(\x|
\thetab)$. The quantity $\phi(\x; \thetab)$ is called an unnormalised
model. This is because the integral
\begin{equation}
  Z(\thetab) =  \int \phi(\x; \thetab) d\x
  \label{eq:Zdef}
\end{equation}
is typically not equal to one for all values of $\thetab$. The integral is a function of $\thetab$ called the partition function. It can be used to formally convert the
unnormalised model $\phi(\x; \thetab)$ into the normalised model
$p(\x; \thetab)$ via
\begin{align}
  p(\x| \thetab) &= \frac{\phi(\x; \thetab)}{Z(\thetab)}.
\end{align}
However, this relationship is only a formal one since the integral
defining the partition function $Z(\thetab)$ can typically not be
solved analytically in closed form and deterministic numerical
integration becomes quickly infeasible as the dimensionality of $\x$
grows (e.g.\ four or five dimensions are often computationally too
costly already).

EBMs make modelling simpler because specifying a parametrised energy
function $E(\x; \thetab)$ is often much simpler than specifying
$p(\x|\thetab)$. The reason is that we indeed do not need to be
concerned with the normalisation condition that $\int p(\x|\thetab)
d\x =1$ for all values of $\thetab$. This enables us, for instance, to
use deep neural networks to specify $E(\x; \thetab)$. The flip side of
this relaxation in the modelling constraint is that $Z(\thetab)$ can
generally not be computed, which means that $p(\x| \thetab)$ and the
likelihood function $L(\thetab)$ are computationally intractable. An
additional issue with energy-based models is that sampling from them typically
requires MCMC methods, which is difficult to scale to the data dimensions of
e.g.\ text or images \citep{Nijkamp2020, Grathwol2021}.

\begin{figure*}
  \centering
  \includegraphics[width=\textwidth]{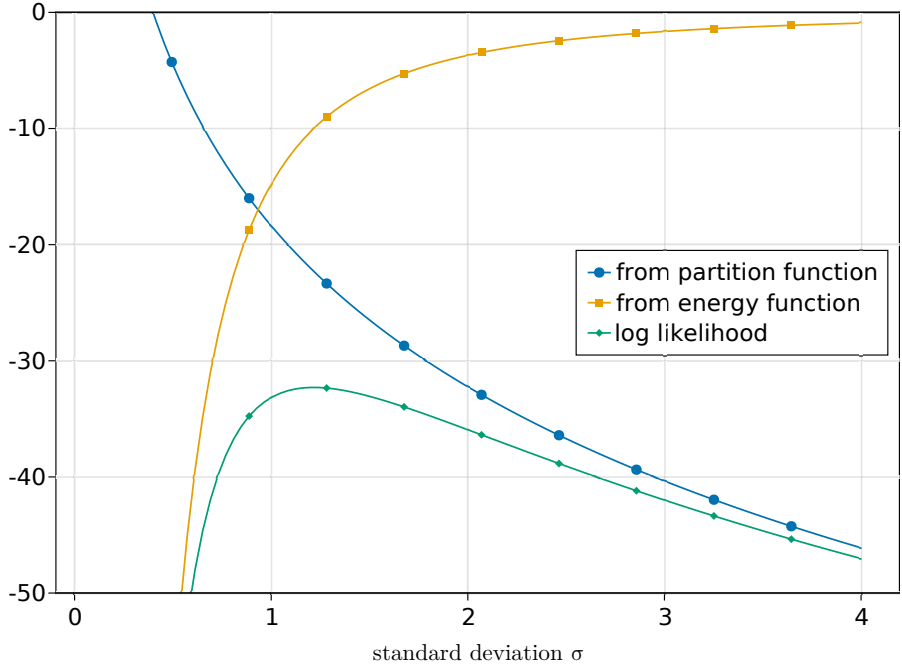}
\caption{\label{fig:loglikdecomp}The log-likelihood function
  $\ell(\sigma)$ has two components that balance each other: the term
  due to the partition function monotonically decreases while the term
  due to the energy function monotonically increases. The balance
  between the two terms leads to a likelihood function $\ell(\sigma)$
  with a well-defined optimum.}
\end{figure*}

To see the importance of the partition function, consider the simple
toy example
\begin{align}
  p(x; \sigma) & = \frac{\exp(-E(x; \sigma))}{Z(\sigma)},& E(x; \sigma)
    &= \frac{x^2}{2\sigma^2}
\end{align}
with $x \in \mathbb{R}$ and parameter $\sigma > 0$. This
corresponds to an unnormalised Gaussian model. The partition function
here is well-known and equals
\begin{equation}
  Z(\sigma) = \sqrt{2\pi \sigma^2}.
\end{equation}
For $n$ data points $\{x_1, \ldots, x_n\}$, the log-likelihood $\ell(\sigma)$ thus is
\begin{align}
  \ell(\sigma) &= \log \prod_{i=1}^n \frac{\exp(-E(x_i;\sigma))}{Z(\sigma)}\\
  & = - n \log Z(\sigma) - \sum_{i=1}^n E(x_i; \sigma) \\
  & = \underbrace{-\frac{n}{2} \log(2\pi \sigma^2)}_{\text{from partition function}} - \underbrace{\frac{1}{2\sigma^2} \sum_{i=1}^n x_i^2}_{\text{from the energy}}.
\end{align}
Note that the term from the partition function does not depend on the
data but on the parameter $\sigma$, which crucially contributes to
the log-likelihood function. It monotonically decreases as $\sigma$
increases while the data-dependent term from the energy
monotonically increases. This leads to a log-likelihood function with
a well-defined optimum as illustrated in Figure
\ref{fig:loglikdecomp}.

The contribution of the (log) partition function on the (log)
likelihood function has two consequences: First, we cannot simply
ignore the partition function if it is difficult to compute. In the
simple Gaussian example in Figure \ref{fig:loglikdecomp}, the maximum
of the term due to the energy is achieved for $\sigma \to \infty$
irrespective of the data, which is not a meaningful
estimator. Secondly, if we approximate the partition function,
possible errors in the approximation may shift the location of the
maximum, and hence affect the quality of the estimate.

\subsection{Simulator-based model}
\label{sec:SBM}
Simulator-based models (SBMs) are another widely used class of
models. They occur in multiple fields, for instance genetics
\citep[e.g.][]{Beaumont2002, Marttinen2015}, epidemiology
\citep[e.g.][]{Allen2017, Parisi2021}, systems biology
\citep[e.g.][]{Wilkinson2018}, cosmology \citep[e.g.][]{Schafer2012} or econometrics \citep[e.g.][]{Gourieroux1996}, just to name a few. Different research communities use different names so that SBMs are also known as e.g.\ implicit models \citep{Diggle1984} or stochastic simulation models \citep{Hartig2011}.

SBMs are specified via a measurable function $g$ that is typically not
known in closed form but implemented as a computer programme. The
function $g$ maps parameters $\thetab$ and realisations of some base
random variable $\omegab$ to data $\x$, i.e.\
\begin{align}
  \x &= g(\omegab, \thetab), & \omegab &\sim p(\omegab),
  \label{eq:SBM}
\end{align}
where $p(\omegab)$ denotes the pdf of $\omegab$. Without loss of
generality, we can assume that $\omegab$ are a collection of independent random
variables uniformly distributed on the unit interval.

\begin{figure*}
  \centering
  \includegraphics[width=0.9\textwidth]{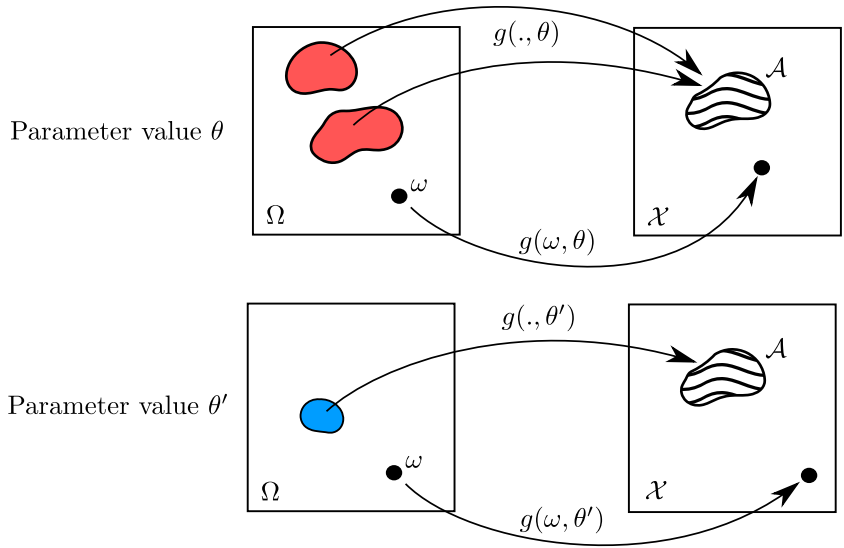}
  \caption{\label{fig:SBMproba} The distribution of $\x$ in a
    simulator-based model is defined by the inverse-image of the
    intractable mapping $g(\omegab, \thetab)$ that maps base random
    variables $\omegab$ to data $\x$, and the inverse image changes
    when $\thetab$ is varied. The mapping $g$ is typically not known
    in closed form but implemented as a computer programme. This setup
    makes evaluating $p(\x|\thetab)$ typically impossible.}
\end{figure*}

The distribution of $\x$ is defined by $g$ and the distribution of
$\omegab$: The probability that $\x$ takes on some values in a set
$\mathcal{A}$ is defined as
\begin{equation}
  \Pr(\x \in \mathcal{A}|\thetab) = \Pr( \{\omega: g(\omegab, \thetab)\in \mathcal{A}\}),
\end{equation}
where the probability on the right-hand side is computed with respect
to the distribution of $\omegab$. The randomness of $\omegab$ implies
the randomness on the level of $\x$ and the function $g$ is thus said to
``push forward'' the distribution of $\omegab$ to $\x$. Determining the
set of all $\omegab$ that are mapped to $\mathcal{A}$ for a given
$\thetab$ corresponds to the problem of determining the inverse-image
of $\mathcal{A}$ under $g$, see Figure \ref{fig:SBMproba}. 

Whilst the distribution of $\x$ is well defined for a given value of
$\thetab$ (assuming that some technical measure-theoretical conditions on $g$ hold), writing
down or computing the conditional pdf $p(\x | \thetab)$ given the pdf
$p(\omegab)$ becomes quickly problematic when $g$ is not
invertible. Moreover, in case of simulator-based models, we typically
do not know a closed-form expression for $g$, which
makes computing, or exactly evaluating, $p(\x | \thetab)$ impossible.

The main advantage of SBMs is that they neatly connect the natural
sciences with statistics and machine learning. We can use principled
tools from statistics and machine learning to perform inference and
experimental design for models from the natural sciences. The flip side, however, is that the likelihood function---the key workhorse for
inference and experimental design---is typically intractable because
the conditional pdf $p(\x | \thetab)$ is intractable.

\section{Contrastive learning}
\label{sec:contrastive-learning}
Contrastive learning is an alternative to likelihood-based learning
when the model pdf and hence likelihood function is intractable. We
first explain the basic idea and then discuss the two main ingredients
of contrastive learning: the loss function and the choice of the
contrast, or reference data.

\subsection{Basic idea}
\label{sec:basic-idea}
The basic idea in contrastive learning is to learn the difference
between the data of interest and some reference data. The properties
of the reference are typically known or not of interest; by learning
the difference we thus focus the (computational) resources on learning
what matters.

Assuming that we are interested in a quantity $a$ and that our
reference is $b$, we can deduce $a$ from $b$ when we know the
difference between them:
\begin{equation}
  \underbrace{b}_{\text{reference}} + \underbrace{a-b}_{\text{difference}} \Rightarrow \underbrace{a}_{\text{interest}}
\end{equation}
This straightforward equation captures much of the essence of
contrastive learning. This means that if we have some reference $b$
available, we can learn about $a$ by contrasting $a$ with $b$ rather than
by starting from scratch.

There is an immediate link to (log) ratio estimation
\citep[e.g.][]{Sugiyama2012} when we let the quantity of interest and
reference be $\log p_a$ and $\log p_b$, respectively,
\begin{equation}
      \underbrace{\log p_b}_{\text{reference}} + \underbrace{\log p_a-
        \log p_b}_{\text{difference}} \Rightarrow \underbrace{\log
        p_a}_{\text{interest}}.
      \label{eq:contrastive-learning-log-ratio}
 \end{equation}
Here, contrasting $a$ with $b$ means learning the log-ratio $\log
p_a-\log p_b$. This connects to logistic regression, and hence
classification, which we will heavily exploit in this paper. The
connection to classification makes intuitive sense since learning the
difference between data sets is indeed what we need to do when solving
a classification problem.

Denoting the log-ratio $\log p_a-\log p_b$ by $h$, let us write the above equation as
\begin{equation}
  p_a = \exp(h) p_b.
  \label{eq:change-of-measure}
\end{equation}
We can thus express $p_a$ as a change of measure from $p_b$
where the learned log-ratio (difference) $h$ determines the change of
measure.

There is also a direct connection to Bayesian inference because Bayes' rule
\begin{equation}
  p(\thetab | \x )  = \frac{p(\thetab, \x)}{p(\x)} = \frac{p(\x | \thetab)p(\thetab)}{p(\x)}
\end{equation}
can be rewritten in the style of \eqref{eq:contrastive-learning-log-ratio} as
\begin{equation}
 \underbrace{\log p(\thetab)}_{\text{reference}} + \underbrace{\log p(\x|\thetab)-
        \log p(\x)}_{\text{difference}} \Rightarrow \underbrace{\log
        p(\thetab|\x)}_{\text{interest}},
\end{equation}
with the log posterior as our quantity of interest $\log p_b$, the log prior as our reference $\log p_a$, and their difference (the ``contrast'') proportional to the log likelihood. This reflects the role of the likelihood function in Bayesian inference as the quantity that transforms the prior belief into an unnormalised posterior belief.

\subsection{Loss functions}
\label{sec:loss-functions}
In the following, we focus on the logistic, or log, loss as done in
early work on contrastive learning to estimate energy-based models,
i.e.\ noise-contrastive estimation \citep[NCE,][]{Gutmann2010,
  Gutmann2012a}. However, other loss functions can be used,
e.g.\ Bregman divergences \citep{Gutmann2011b, Sugiyama2012b} or
f-divergences \citep{Nowozin2016}, see also \citet{Mohamed2017, pmlr-v97-poole19a}.

Given data of interest $\x_o = \{\x_1, \ldots, \x_n\}$, with $\x_i
\sim p$ (iid), and reference data $\{\y_1, \ldots, \y_m\}$, $\y_i
\sim q$ (iid), let us label the data points as follows. The $\x_i$
become tuples $(\x_i, 1)$ and the $\y_i$ become tuples $(\y_i,
0)$. Learning the difference between the $\x_i$ and $\y_i$ can then be
cast as a problem of predicting the label for test points, which
corresponds to classification.

A popular loss function for classification is the logistic loss, which
means that we perform classification via (nonlinear) logistic
regression. We denote this loss, normalised by the number of data
points $n$, by $J(h)$,
\begin{align}
  J(h) =& \frac{1}{n} \sum_{i=1}^{n} \log\left[1+\nu
    \exp(-h(\x_i))\right] + \frac{\nu}{m}\sum_{i=1}^{m}
  \log\left[1+\frac{1}{\nu} \exp(h(\y_i))\right],
  \label{eq:logistic-loss}
\end{align}
where $\nu = m/n$ and the parameter of the loss is the function $h$. The
minimal loss is achieved for a function $h$ that assigns large
positive numbers to the $\x_i$ and large negative numbers to the
$\y_i$. It can be shown \citep[e.g.][]{Gutmann2012a, Thomas2020}
that for large sample sizes $n$ and $m$ (and fixed ratio $\nu$), the
optimal regression function $h$ equals
\begin{equation}
  h^\ast = \log p - \log q.
  \label{eq:h-logistic-regression}
\end{equation}
This means that by solving the classification problem via logistic
regression we learn the density ratio between $p$ and $q$. This
important result reflects the connection between the different
concepts discussed above, namely contrastive learning, classification
via logistic regression, and density ratio estimation. The result
implies consistency of the estimator for parametric models and finite
amount of data under some technical conditions \citep[e.g.][Chapter
  4]{Amemiya1985}.

Further important properties of the logistic loss are:
\begin{enumerate}
\item The optimal $h^\ast$ in \eqref{eq:h-logistic-regression} is
  ``automagically'' the difference between two log \emph{densities}. This
  holds without us having to specify that $h$ should take this
  particular form. We will exploit this property in the estimation of
  unnormalised models, as well as in the Bayesian inference and
  experimental design for simulator-based models.
  \label{prop1}
\item We only need samples from $p$ and $q$; we do not need their
  densities or models. This is important because we do not need
  to specify properly normalised pdfs as in likelihood-based learning
  and experimental design. However, we do need to model their
  ratio. This can be exploited as a modelling tool because it may
  often be easier to specify how the data of interest differs from the
  reference rather than specifying a parametric family of pdfs
  from scratch.
  \label{prop2}
\end{enumerate}
Both properties will be heavily exploited in the applications of
contrastive learning to statistical inference and experimental
design. Other loss functions mentioned above have the same properties.

When the sample sizes $n$ and $m$ are arbitrarily large, by the law of
large numbers, the sample averages over $\x_i$ and $\y_i$ in
\eqref{eq:logistic-loss} become expectations with respect to the
densities $p$ and $q$, respectively. Denote the corresponding limiting
loss function of $J(h)$ in \eqref{eq:logistic-loss} by
$\bar{J}_\nu(h)$. Below, we will denote the limiting loss function in
case of $\nu=1$ by $\bar{J}(h)$.

It is illustrative to consider the minimal loss $\bar{J}_\nu(h^\ast)$ that is achieved by $h^\ast$,
\begin{align}
  \bar{J}_\nu(h^\ast) & = \E_{\x\sim p} \log \left[1+\nu \frac{q(\x)}{p(\x)}\right] + \nu \E_{\y \sim q} \log \left[1+ \frac{p(\y)}{\nu q(\y)}\right]\\
  &= \E_{\x\sim p} \log \left[\frac{p(\x) + \nu q(\x)}{p(\x)}\right] + \nu \E_{\y \sim q} \log \left[ \frac{\nu q(\y)+p(\y)}{\nu q(\y)}\right]\\
  &= - \E_{\x\sim p} \log \left[\frac{p(\x)}{p(\x) + \nu q(\x)}\right] - \nu \E_{\y \sim q} \log \left[ \frac{\nu q(\y)}{\nu q(\y)+p(\y)}\right].
  \label{eq:Jbar}
\end{align}
Setting $\nu=1$ and introducing the mixture density $m = (p+q)/2$, we
can write the above in terms of two Kullback-Leibler divergences,
which furthermore can be expressed as the Jensen-Shannon divergence (JSD) between $p$ and $q$,
\begin{align}
  \bar{J}(h^\ast) & \overset{\nu=1}{=} - \E_{\x\sim p} \log \left[\frac{p(\x)}{p(\x) + q(\x)}\right] - \E_{\y \sim q} \log \left[ \frac{q(\y)}{q(\y)+p(\y)}\right]   \label{eq:Jbarnu1}\\
  & =  -\text{KL}(p|| m) - \text{KL}(q||m) +2\log 2\\
  & =-\text{JSD}(p, q) + 2 \log 2.
\end{align}
Since $\bar{J}(h) \ge  \bar{J}(h^\ast)$, we have the following bound
\begin{equation}
  \bar{J}(h) \ge  -\text{JSD}(p, q) + 2 \log 2.
  \label{eq:JSD-varbound}
\end{equation}
We can thus think that minimising $\bar{J}(h)$ with respect to $h$
leads to a variational estimate of the negative Jensen-Shannon
divergence between $p$ and $q$ up to a known additive constant, which can be used to quantify the difference between $p$
and $q$ in the constructed classification problem.

\subsection{Reference data}
\label{sec:reference-data}
Much of the creative aspect of contrastive learning lies in the choice
of the reference data. The choice depends on the application of
contrastive learning, e.g.\ whether we are interested in model
estimation or Bayesian inference. While intuition can partly guide the
choice, establishing optimality results is an open research
direction.

A simple option is to fit a preliminary model to the observed data
$\xobs$ and to use data generated from the model as the
reference. This has been done in the first work on estimating
energy-based models with contrastive learning \citep[noise-contrastive
  estimation,][]{Gutmann2012a}.  An extension is to iterate the above
approach so that the fitted model in one iteration becomes the
reference distribution in the next. Simulation results by
\citet{Gutmann2010} showed that this improved estimation accuracy
compared to the ``static'' approach where the reference (noise)
distribution was kept fixed. The reason for this is that at the start
of a new iteration, the reference data includes all the information
of the real data that was captured by the model so far, and a model-based
classifier would not be able to distinguish between the reference data
and the real data. The iteration forces the system to focus on the
aspects of the real data that have not yet been captured by the model.

A further option is to create the reference data conditional on each observed data point so that each reference data point is paired with an observed data point. This is
useful if the observed data lies on a lower-dimensional manifold for
which fitting a preliminary model may be difficult \citep{Ceylan2018}.

In the iterative scheme above, each iteration essentially resets the
classification performance to a 50\% chance level. However, rather
than periodically resetting the classification performance, one can
also continuously update the reference distribution to push the
classification performance towards chance level. This kind of
iterative adaptive approach, which is known as ``adversarial
training'', has been used to estimate the parameters of a generative
model specified by a neural network \citep[generative adversarial
  networks,][]{Goodfellow2014}. The generative model can be seen to
correspond to a simulator-based model where the function $g$ in
\eqref{eq:SBM} is given by a neural network and the parameters
$\thetab$ that we would like to estimate are its weights. A similar
adaptive scheme has been proposed by \citet{Gutmann2014, Gutmann2018}
for both point and posterior estimation of general simulator-based
models, see also the review by \citet{Mohamed2017}. 

A complementary perspective is given by the variational bound in
\eqref{eq:JSD-varbound}: learning the nonlinear regression function
$h$ provides us with an estimate of the Jensen-Shannon divergence
between the data distribution and the reference, and the adaptive
updating of the reference distribution $q$ corresponds to a further
optimisation that pushes the reference distribution $q$ towards the
data distribution $p$ so that the Jensen-Shannon divergence between
them is minimised.

\citet{Gao2020} used such an iterative adaptive scheme to learn an
energy-based model choosing as reference distribution a normalising
flow \citep[see][for a comprehensive introduction to normalising flows]{Papamakarios2021a}. The normalising flow is continuously
updated with the learning of the energy-based model to provide a
strong reference distribution. As in case of the generative
adversarial network above, the complementary perspective is that the
flow is learned by (approximately) minimising the Jensen-Shannon to
the data distribution. The benefit of this approach is twofold: On the
one hand, the method enables the learning of an energy-based model in
high-dimensions and is well suited for semi-supervised learning. On
the other hand, the learned flow is by itself of interest since (a) it
allows for exact sampling and (b) it is normalised and thus allows for
likelihood evaluations.

When modelling time-series data, \citet{Hyvarinen2016} proposed to use
windows of the time-series itself as the reference data. Under certain
assumptions, this approach was shown to enable the estimation of
independent components in a nonlinear time-series model
\citep{Hyvarinen2016}.

For Bayesian inference and experimental design, a natural choice of
the reference data is to use the prior predictive distribution in line
with the intuition in Section \ref{sec:basic-idea}. This can be used
to estimate the posterior distribution for simulator-based models
and to perform experimental design as we will discuss below.
\begin{figure*}
  \centering
  \includegraphics[width=\textwidth]{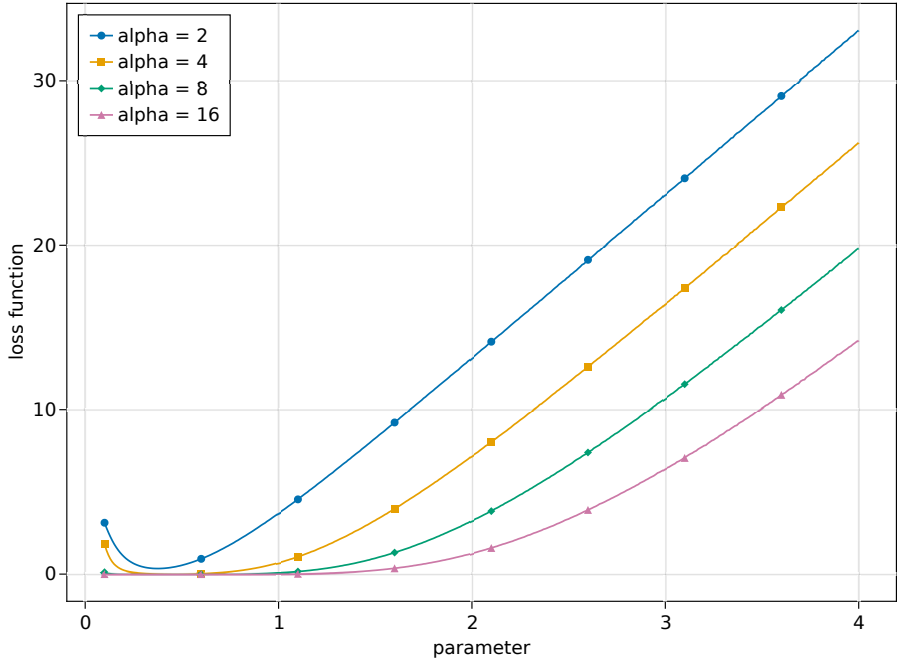}
\caption{\label{fig:chasm}Logistic loss to learn the difference
  between a standard normal distribution and a Gaussian with standard
  deviation $\alpha$ in 10 dimensions. The log-ratio was parametrised
  by one free parameter acting on the squared norm of the random
  variables. The loss function becomes flatter around the optimum as
  $\alpha$ increases, i.e.\ as the two distributions become more different.}
\end{figure*}

A major difficulty in contrastive learning with logistic regression is
that learning the difference between the reference and data
distribution accurately is difficult if they are very
different. Intuitively, highly different distributions correspond to
easy classification problems so that the classifier does not need to
learn a lot about the structure of the data sets to achieve good
performance. Another view is that several different classification
boundaries can achieve similar performance, so that the loss function
is flat around the optimum, which causes a larger estimation
error. Figure \ref{fig:chasm} illustrates this behaviour in case of
the logistic loss: the data follows a standard normal distribution and
the reference is a Gaussian distribution with standard deviation
$\alpha >1$ (in 10 dimensions). As $\alpha$ becomes larger, the
reference distribution becomes more dissimilar from the data
distribution and the loss function flatter around the optimum even
though the scale of the loss function is about the same (and for large
$\alpha$, the optimal parameter value equals 0.5 for all $\alpha$).

\citet{Rhodes2020a} called this issues the ``density chasm'' problem
and proposed a divide-and-conquer strategy to learn the ratio: Rather
than attempting to learn the difference between $p_a$ and $p_b$ in one
go, they introduce auxiliary distributions that anneal between $p_a$
and $p_b$. The auxiliary distributions are constructed so that the
differences between them are sufficiently small so that their ratio can
be better estimated, and the different estimates are then combined via
a telescoping product (or sum in the log domain), i.e.\
\begin{equation}
  \underbrace{\log p_a - \log p_b}_{\text{large gap}} = \underbrace{\log p_a - \log p_1}_{\text{small gap}} + \sum_{k=1}^{K-1} \underbrace{(\log p_k - \log p_{k+1})}_{\text{small gap}} + \underbrace{\log p_m - \log p_b}_{\text{small gap}},
\end{equation}
where the $p_i$ are the auxiliary densities. The functional form or a
model of the auxiliary distributions does not need to be known since
the method only requires samples from them. It is further possible to
derive a limiting objective function when the number of auxiliary
distributions $K$ goes to infinity, which was shown to lead to
improved performance and removes the need to choose $K$
\citep{Choi2021}.

Telescoping density-ratio estimation by \citet{Rhodes2020a, Choi2021}
deals with the density chasm and the flat loss landscape by
re-formulating the contrastive problem. A complementary algorithmic
approach was taken by \citet{Liu2021a} who asked whether optimisation
techniques can deal with the flat optimisation landscape of the
logistic loss when there is a density chasm. They found that, in case
of exponential families, normalised gradient descent can deal with the
issue. Moreover, they showed that a polynomial convergence guarantee
can be obtained when working with the exponential rather than the
logistic loss. The exponential loss is, like the logistic loss, a
Bregman divergence that was shown to provide a large family of loss
functions for the contrastive learning of energy-based models
\citep{Gutmann2011b}. The result by \citet{Liu2021a} highlights that
for some models, particular instances of the family of loss functions
are more suitable than others.

\section{Applications in statistical inference and experimental design}
\label{sec:applications}
We here show how we can use contrastive learning to estimate energy-based
models and perform Bayesian inference and experimental design for
simulator-based models. 

\subsection{Estimating energy-based models}
We introduced energy-based models (EBMs) in Section \ref{sec:EBM},
pointing out that they are unnormalised models $\phi(\x; \thetab) =
\exp(-E(\x; \thetab))$ that are specified by an energy function $E(\x;
\thetab)$. We have seen that standard maximum likelihood
estimation cannot be used to learn the parameters $\thetab$ if the
partition function $Z(\thetab)$ in \eqref{eq:Zdef} is not available
analytically in closed form.

Contrastive learning can be used to estimate energy-based models from
data $\xobs = (\x_1, \ldots, \x_n)$, $\x_i \sim p$ (iid) by
introducing reference data $(\y_1, \ldots, \y_m)$, $\y_i \sim q$ (iid)
and estimating the log-ratio $h^\ast = \log p - \log q$. An estimate
$\hat{h}$ thus provides an estimate of $p$ when $q$ is known,
\begin{equation}
  \hat{p} = \exp(\hat{h}) q,
  \label{eq:EBMhat}
\end{equation}
where the factor $\exp(\hat{h})$ is a form of exponential tilting
or a change of measure from $q$ to $p$ in line with
\eqref{eq:change-of-measure}.

There are different ways to parametrise $h$: If we would like to
estimate an energy-based model where $\phi(\x; \thetab)$ has a
specific form, we parametrise $h$ as
\begin{equation}
  h(\x; \thetab, c) = \log \phi(\x;\thetab) - \log q(\x) + c.
  \label{eq:hparam-EBM}
\end{equation}
From \eqref{eq:EBMhat}, we can see that $q$ cancels out and hence
\begin{equation}
  \hat{p}(\x) = \exp(\hat{c}) \phi(\x; \hat{\thetab}).
\end{equation}
The parameter $c$ allows for scaling of $\phi(\x; \thetab)$ and can optionally be
included if $\phi(\x; \thetab)$ is not flexible enough. On the other
hand, if we are not interested in estimating a specific model of a
particular form, we may also parametrise $h$ directly, e.g.\ as a deep neural network in unsupervised deep learning.

This principle to learn energy-based models was proposed by
\citet{Gutmann2010, Gutmann2012a} using the logistic loss. They called
the reference distribution ``noise'' and the estimation principle
hence ``noise-contrastive-estimation''. The term ``noise'' should not
be mis-understood to refer to unstructured data. As explained in
Section \ref{sec:contrastive-learning}, it can be structured and
reflect our current understanding of the properties of the observed
data $\xobs$. The above equations highlight two key conditions for the
reference (noise) distribution $q$: (1) we need to be able to sample
from it, (2) we need to know an analytical expression for it (strictly
speaking, up to the normalisation constant only).

The fact that the model does not need to be normalised is due to
property \ref{prop1} of the logistic loss, namely that it yields the
difference between two log densities, as in
\eqref{eq:h-logistic-regression}, without any normalisation
constraints. Further loss functions with this property were proposed
by \citet{Pihlaja2010, Gutmann2011b, Uehara2020a}. Moreover, the
telescoping approach by \citet{Rhodes2020a, Choi2021} can also be used
to learn $h$, improving upon the single-ratio methods. For an overview
of further methods to estimate energy-based models, we refer the reader to the
recent review by \citet{Song2021}, and for the case of energy-based
models with latent (unobserved) variables to the work by \citet{Rhodes2019}.

In Section \ref{sec:reference-data}, we discussed an iterative
approach where the estimated model from a previous iteration is used
as reference \citep{Gutmann2010}. \citet{Goodfellow2014b} related this
approach to maximum likelihood estimation. We here provide a
simplified proof of the relationship. Let the log-ratio $h$ be
parametrised by $\thetab$ and consider the loss function in
\eqref{eq:logistic-loss} as function of $\thetab$,
\begin{align}
  J(\thetab) &= \frac{1}{n} \sum_{i=1}^{n} \log\left[1+\nu
    \exp(-h(\x_i; \thetab))\right] + \frac{\nu}{m}\sum_{i=1}^{m}
  \log\left[1+\frac{1}{\nu} \exp(h(\y_i; \thetab))\right],
\end{align}
with  $\y_i \sim q$ (iid) and $\nu = m/n$. The gradient with respect to $\thetab$ is
\begin{align}
  \nabla_{\thetab}  J(\thetab) =&  \frac{1}{n} \sum_{i=1}^{n} \frac{-\nu \exp(-h(\x_i; \thetab))}{1+\nu
    \exp(-h(\x_i; \thetab))}\nabla_{\thetab}h(\x_i; \thetab) + \nonumber\\
  &\frac{1}{m}\sum_{i=1}^{m} \frac{\exp(h(\y_i; \thetab))}{1+\frac{1}{\nu} \exp(h(\y_i; \thetab))}\nabla_{\thetab}h(\y_i; \thetab).
  \label{eq:NCE-grad}
\end{align}
Assume now that the reference data $\y$ follows the distribution of the
model at iteration $t$, i.e.\ $p(.;\thetab_t)$. For a parametrisation
of $h$ as in \eqref{eq:hparam-EBM}, we then have
\begin{equation}
  h(\x; \thetab, c) = \log \phi(\x;\thetab) - \log p(\x;\thetab_t) +c
  \label{eq:hparam-iterative}
\end{equation}
and the $\y_i \sim p(.;\thetab_t)$. The gradient of $h(\x; \thetab,
c)$ with respect to $\thetab$ is
\begin{equation}
  \nabla_{\thetab} h(\x; \thetab, c) =  \nabla_{\thetab} \log \phi(\x;\thetab)
\end{equation}
and hence
\begin{align}
  \nabla_{\thetab}  J(\thetab) =&  \frac{1}{n} \sum_{i=1}^{n} \frac{-\nu \exp(-h(\x_i; \thetab))}{1+\nu
    \exp(-h(\x_i; \thetab))} \nabla_{\thetab} \log \phi(\x_i;\thetab) + \nonumber\\
  &  \frac{1}{m}\sum_{i=1}^{m} \frac{\exp(h(\y_i; \thetab))}{1+\frac{1}{\nu} \exp(h(\y_i; \thetab))}\nabla_{\thetab} \log \phi(\y_i;\thetab)
\end{align}
with  $\y_i \sim p(.;\thetab_t)$. Let us update $\thetab$ by gradient descent on $J(\thetab)$ via
\begin{equation}
  \thetab_{t+1} = \thetab_t - \epsilon  \nabla_{\thetab} J(\thetab)\big|_{\thetab=\thetab_t},
\end{equation}
where $\epsilon$ is a small step-size. Note that in the evaluation of
the gradient at $\thetab_t$, with \eqref{eq:hparam-iterative},
$h(\x_i; \thetab_t, c) = h(\y_i; \thetab_t, c) = c+\log Z(\thetab_t)$,
which is independent of $\x_i$ and $\y_i$. Let us denote its value by
$b_t$.  It measures the error in the normalisation of the unnormalised
model: $b_t$ is close to zero if the parameter $c$ is close to the
logarithm of the inverse partition function at $\thetab_t$,
e.g.\ towards the end of the optimisation. Moreover, $b_t$ would be
always zero in the special case of normalised models.

We thus obtain
\begin{align}
  \frac{-\nu \exp(-h(\x_i; \thetab))}{1+\nu \exp(-h(\x_i; \thetab))} & =  \frac{-\nu \exp(-b_t)}{1+\nu \exp(-b_t)}\\
  \frac{\exp(h(\y_i; \thetab))}{1+\frac{1}{\nu} \exp(h(\y_i; \thetab))} & =  \frac{\nu}{1+\nu \exp(-b_t)}
\end{align}
and hence
\begin{align}
  \nabla_{\thetab} J(\thetab)\big|_{\thetab=\thetab_t} =& \frac{-\nu}{1+\nu
    e^{-b_t}}\left(\frac{e^{-b_t}}{n} \sum_{i=1}^{n}
  \nabla_{\thetab} \log \phi(\x_i;\thetab) - \frac{1}{m} \sum_{i=1}^m
  \nabla_{\thetab} \log \phi(\y_i;\thetab)\right)\Big|_{\thetab=\thetab_t}
  \label{eq:EBM-grad}
\end{align}
On the other
hand, the gradient of the negative log-likelihood is
\begin{align}
  -\nabla_{\thetab} \ell(\thetab)\big|_{\thetab=\thetab_t} =& -\left(\frac{1}{n}
  \sum_{i=1}^{n} \nabla_{\thetab} \log
  \phi(\x_i;\thetab) -\E_{p(\x;\thetab_t)}\left[  \nabla_{\thetab} \log \phi(\x_i;\thetab)\right]\right)\Big|_{\thetab=\thetab_t}
  \label{eq:ell-grad}
\end{align}
The average over the $\y_i \sim p(.; \thetab_t)$ in
\eqref{eq:EBM-grad} is a Monte Carlo estimate of the derivative of the
log-partition function, which is the second term in
\eqref{eq:ell-grad}.  Hence for $b_t=0$, e.g.\ in case of normalised
models, the gradient for contrastive learning with the logistic loss
and $p(.; \thetab_t)$ as reference distribution is a Monte Carlo
approximation of the gradient of the negative log-likelihood (up to
constant scaling factor that can be absorbed into the
step-size). There is thus a clear connection between contrastive
learning and maximum likelihood estimation: if, at every gradient
step, we use the current model $p(.; \thetab_t)$ as reference
distribution, we follow noisy gradients of the (negative)
log-likelihood.

This scheme, however, is typically computationally not
feasible since sampling from $p(.; \thetab_t)$ is prohibitively
expensive. But we are not required to use at \emph{every}
iteration the current model. In contrastive learning, we are allowed
to use the same ``old'' $p(.; \thetab_t)$ to update the parameter
$\thetab$ in the subsequent iterations. This results in a valid
estimator (as in standard noise-contrastive estimation with a fixed
reference distribution), but the gradient updates would not correspond
to Monte Carlo approximations of the gradient of the negative
log-likelihood.

A possibly simpler connection to maximum likelihood estimation can be
obtained by considering the case of large $\nu$. \citet{Gutmann2012a}
showed that for normalised models, noise-contrastive estimation is
asymptotically equivalent to maximum likelihood estimation in the
sense that the estimator has the same distribution irrespective of the
choice of the reference distribution $q$ (as long as it satisfies some
weak technical conditions). Moreover, \citet{Mnih2013b} showed that
the gradient converges to the gradient of the negative log-likelihood.
We can see this from \eqref{eq:NCE-grad} for $\nu \to \infty$: With
$h(.; \thetab) = \log \phi(.; \thetab) - \log q(.)$, and
\begin{align}
  \lim_{\nu \to \infty}  \frac{-\nu \exp(-h(\x; \thetab))}{1+\nu \exp(-h(\x; \thetab))} & = -1\\
  \lim_{\nu \to \infty}  \frac{\exp(h(\y; \thetab))}{1+\frac{1}{\nu} \exp(h(\y; \thetab))} & = \exp(h(\y; \thetab)) = \frac{\phi(\y; \thetab)}{q(\y)},
\end{align}
taking the limit of $\nu \to \infty$ in \eqref{eq:NCE-grad} thus gives
\begin{align}
\lim_{\nu \to \infty} \nabla_{\thetab} J(\thetab) =& -\frac{1}{n}
\sum_{i=1}^{n} \nabla_{\thetab}\log \phi(\x_i; \thetab) + \E_{q(\y)}
\left[ \frac{p(\x; \thetab)}{q(\y)} \nabla_{\thetab}\log \phi(\y; \thetab)\right]\\
& =  -\frac{1}{n}
\sum_{i=1}^{n} \nabla_{\thetab}\log \phi(\x_i; \thetab) + \E_{p(\x; \thetab)}
\left[ \nabla_{\thetab}\log \phi(\y; \thetab)\right],
\end{align}
where where have used that for $\nu \to \infty$, $m = \nu n$ becomes
arbitrarily large too, so that the average over the $\y_i$ becomes an
expectation with respect to $q(\y)$. A more general result was
established by \citet{RiouDurand2018}. Moreover, they further
considered the case of finite $\nu$, but large sample sizes $n$, and
showed that the variance of the noise-contrastive estimator is always
smaller than the variance of the Monte Carlo MLE estimator
\citep{Geyer1994} where the partition function is approximated with a
sample average, assuming in both cases that the auxiliary/reference
distributions were fixed.

\subsection{Bayesian inference for simulator-based models}
\label{sec:LFI}
Simulator-based models are specified by a stochastic simulator that is
parametrised by a parameter $\thetab$. While we can generate (sample)
data $\x$ by running the simulator, the conditional pdf $p(\x
|\thetab)$ of the generated data is typically not known in closed
form, see Section \ref{sec:SBM}. This makes standard likelihood-based
learning of the parameters $\thetab$ impossible. The problem of
estimating plausible values of $\thetab$ from some observed data
$\xobs$ is called approximate Bayesian computation, likelihood-free
inference, or simulator/simulation-based inference depending on the
research community. For an introduction to the field, see the review
papers by \citet{Lintusaari2017, Sisson2018, Cranmer2020}.

We related contrastive learning to Bayes' rule in Section
\ref{sec:basic-idea}. This connection is not merely conceptual but can
be turned into an inference method to estimate the posterior pdf of
the parameters $\thetab$ given observed data $\xobs$. Starting from
Bayes' rule in the log-domain,
\begin{equation}
  \log p(\thetab|\x) = \log p(\thetab) + \log p(\x|\thetab)-\log p(\x)
\end{equation}
we can estimate $p(\thetab|\x)$ by learning the difference (log-ratio)
$\log p(\x|\thetab)-\log p(\x) = \log p(\x, \thetab)-\log[ p(\x)
  p(\thetab) ]$ by contrastive learning and then combining it with the
prior. To learn the difference, we can exploit that the model is
specified in terms of a simulator that generates data from
$p(\x|\thetab)$, and hence also from the marginal (prior predictive)
pdf $p(\x) = \int p(\thetab) p(\x|\thetab) d \thetab$, which provides
all the data that we need to learn the log-ratio. This approach to
estimate the posterior was introduced by \citet{Thomas2016,
  Thomas2020} and called ``likelihood-free inference by ratio
estimation''. An interesting property of this approach is that the
learning of $h$ can be performed offline, prior to seeing the observed
data $\xobs$, which enables amortisation of the inference.

The logistic loss in \eqref{eq:logistic-loss} is among the several
loss functions that can be used for contrastive learning of the
posterior. A simple approach is to minimise
\begin{align}
  J(h) =& \frac{1}{n} \sum_{i=1}^{n} \log\left[1+\nu
    \exp(-h(\x_i))\right] + \frac{\nu}{m}\sum_{i=1}^{m}
  \log\left[1+\frac{1}{\nu} \exp(h(\y_i))\right]
\end{align}
with $\x_i \sim p(\x|\thetab)$ (iid) and $\y_i \sim q$ (iid) with the
contrastive distribution $q$ being equal to the marginal $p(\x)$. Due to
\eqref{eq:h-logistic-regression}, the optimal $h$ is then indeed the
desired $\log p(\x|\thetab)-\log p(\x)$ for any value of
$\thetab$. Note that we are here again exploiting properties
\ref{prop1} and \ref{prop2} of the logistic loss highlighted in
Section \ref{sec:loss-functions}. 

Since the optimal $h$ is given by $\log p(\x|\thetab)-\log p(\x)$ for
any $\thetab$, we can further amortise the inference with respect to
$\thetab$ by averaging the loss function over different values of
$\thetab$. When averaging over samples from the prior $p(\thetab)$,
this amounts to contrasting samples $(\x, \thetab) \sim
p(\x|\thetab)p(\thetab)$ with samples $(\x, \thetab) \sim p(\x)
p(\thetab)$, which was used by \citet{Hermans2020} together with MCMC to perform Bayesian inference.

In the approach described above, we
contrasted simulated data with simulated data in order to essentially
learn the intractable model pdf $p(\x|\thetab)$. However, it is also
possible to contrast the simulated data with the observed data to
infer plausible values of $\thetab$ \citep{Gutmann2014,
  Gutmann2018}. For further information and connections to other work, including the
learning of summary statistics, we refer the reader to
\citep{Pham2014, Thomas2016, Thomas2020, Dinev2018, Hermans2020,
  Durkan2020, Chen2021a}.

\subsection{Bayesian experimental design for simulator-based models}
Let us now consider Bayesian experimental design for simulator-based
models where the utility of an experiment is measured by the mutual
information (MI) between the data and the quantity of interest \citep{Chaloner1995, Ryan2016}. For
simplicity, we here consider the situation where we are interested in
the parameters $\thetab$ of a model. For other cases such as model
discrimination, see e.g.\ \citep{Ryan2016, Kleinegesse2021b}.

With \eqref{eq:argmaxMI}, we thus would like to solve the following
optimisation problem
\begin{align}
  \hat{\d} &=\argmax_{\d}  \MI_{\d}(\x, \thetab), &  \MI_{\d}(\x, \thetab)&=\E_{p(\x, \thetab|\d)} \log \left[\frac{p(\x |\thetab, \d)}{p(\x|\d)}\right],
  \label{eq:argmaxMI2}
\end{align}
where $\d$ denotes the vector of design parameters. Since we are
dealing with simulator-based models, the densities in the log-ratio
are not available in closed form. This is the same issue as in case of
Bayesian inference for simulator-based models considered above. But
the problem is exacerbated by the following two further issues:
\begin{enumerate}
  \item We need to know or evaluate the log ratio $\log p(\x|\thetab, \d) -
    \log p(\x|\d)$ for (theoretically) infinitely many $\x$ due to the
    expected value in \eqref{eq:argmaxMI2}, and not only the observed
    data $\xobs$ as in Bayesian inference.
  \item Furthermore, we not only need to evaluate the expected value
    of $\log p(\x|\thetab, \d) - \log p(\x|\d)$ but also maximise it with
    respect to $\d$. This is typically done iteratively, which means that the
    log-ratios and expected values need to to re-estimated as $\d$
    changes.
\end{enumerate}
These issues make experimental design for simulator-based models by
mutual information maximisation a highly difficult problem.

\citet{Kleinegesse2019} proposed to build on the properties of the
likelihood-free inference by ratio estimation framework \citep{Thomas2016, Thomas2020} to deal with the issues. As discussed in
Section \ref{sec:LFI}, the method yields an amortised log-ratio
$\log p(\x|\thetab, \d) - \log p(\x|\d)$ where, depending on the setup, the
amortisation is with respect to $\x$ and $\thetab$, but not $\d$, which
allowed them to estimate the expected value in
\eqref{eq:argmaxMI2} via a sample average over the learned
log-ratios. This approach partly deals with the first issue above. For
the optimisation (issue 2), \citet{Kleinegesse2019} used Bayesian
optimisation \citep[e.g.][]{Shahriari2016}, which enables
gradient-free optimisation and smoothes out Monte Carlo errors due to
the sample average.

The work by \citet{Kleinegesse2019} considered the static setting
where prior to the experiment, we would determine the complete
design. The methodology was extended to the sequential setting where
we have the possibility to update our belief about $\thetab$ as we
sequentially acquire the experimental data, see
\citep{Kleinegesse2021a} for further information.

Learning the log ratios and accurately approximating the MI is
computationally costly. However, we do not actually need to estimate
the MI accurately everywhere. First, we are interested in the argument
$\hat{\d}$ that maximises the MI rather than its value. Secondly, it
is sufficient to be accurate around the optimum; cheaper but
noisy or biased estimates of the MI and its gradient are sufficient during the search as long they lead us to $\hat{\d}$.

Such an approach was taken by \citet{Kleinegesse2020a,
  Kleinegesse2021b} who concurrently tightened a variational lower
bound on the mutual information (or proxy quantities) and maximised
the (proxy) MI. Important related work is \citep{Foster2019,
  Foster2020a}. This approach to experimental design for
simulator-based models achieves large computational gains because we
avoid approximating the mutual information accurately for values of
$\d$ away from $\hat{\d}$. \citet{Kleinegesse2021b} showed that a
large class off variational bounds are applicable and can be used for
experimental design with various goal, e.g. parameter estimation or
model discrimination.

Let us consider a special case of the framework that uses contrastive learning via the
logistic loss and the bound in \eqref{eq:JSD-varbound}. Reordering the
terms gives the following lower bound on the Jensen-Shannon divergence
between the two densities $p$ and $q$,
\begin{equation}
  \text{JSD}(p, q)\ge  -\bar{J}(h) + 2 \log 2,
  \label{eq:JSD-lower-bound}
\end{equation}
where $\bar{J}(h)$ is the logistic loss in \eqref{eq:logistic-loss}
for $\nu=1$ and $n, m \to \infty$.  Whilst the mutual information
between data and parameters is defined in terms of the Kullback-Leibler
(KL) divergence between the $p(\thetab, \x | \d)$ and $p(\thetab |
\d)p(\x|\d)$, see \eqref{eq:argmaxMI}, the Jensen-Shannon divergence
has been found to be a practical alternative to the KL divergence due
to its increased robustness \citep[e.g.][]{Hjelm2018}.

\begin{figure*}
\includegraphics[width=\textwidth]{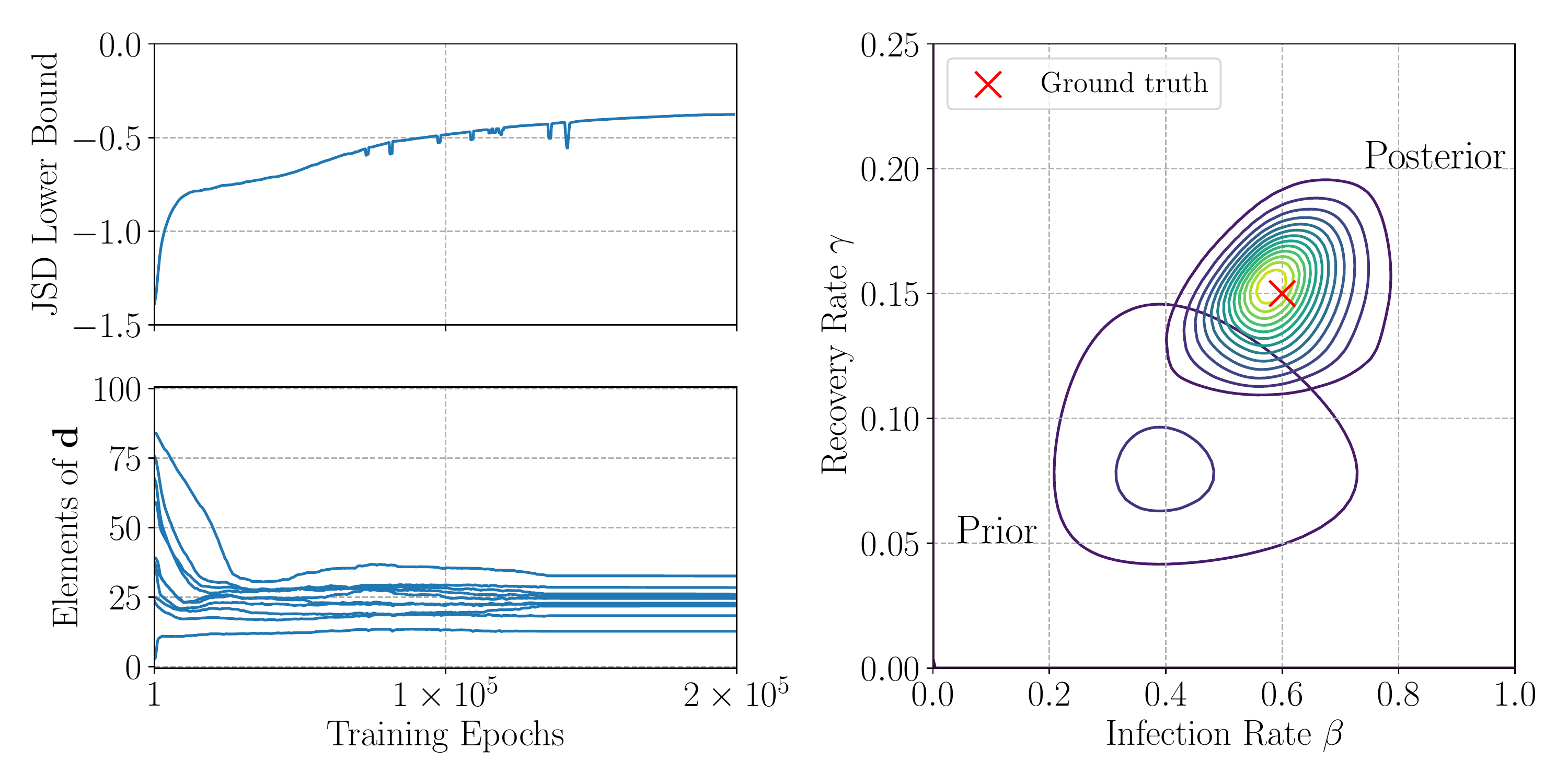}
\caption{\label{fig:BOED}Experimental design and posterior inference by maximising a lower bound on the Jensen-Shannon divergence (JSD) via the logistic loss. The example considers the design of measurement points in a stochastic SIR model in epidemiology. The quantities of interest are the infection rate and the recovery rate. Left top row: optimisation of the JSD lower bound. Left bottom row: learning trajectories of the optimal designs. Right: prior and estimated posterior for data acquired at the optimal design. The data was simulated with the ground-truth parameters indicated with a red cross. See \citet{Kleinegesse2021b} for details.}
\end{figure*}

The lower bound in \eqref{eq:JSD-lower-bound} allows us to perform
experimental design for simulator-based models by letting $p(\thetab,
\x | \d)$ and $p(\thetab | \d)p(\x|\d)$ play the role of $p$ and $q$,
which means that $\bar{J}(h)$ becomes a function of $h$ and implicitly
also of $\d$. Thus maximising $-\bar{J}(h)$, or minimising the
logistic loss $\bar{J}(h)$, jointly with respect to $h$ and $\d$
allows us to concurrently tighten the bound and find the optimal
design $\hat{\d}$. The optimisation of the logistic loss with respect
to $h$ essentially corresponds to likelihood-free inference
by ratio estimation that we discussed in Section \ref{sec:LFI}. Thus,
we here obtain not only the optimal design $\hat{\d}$ but also an
amortised estimate of the posterior $p(\thetab|\x, \hat{\d})$, see
\citep{Kleinegesse2020a, Kleinegesse2021b} for further details.

Figure \ref{fig:BOED} illustrates this approach for the design of
experiments to estimate the parameters of a stochastic SIR model from
epidemiology. The model has two parameters: the rate $\beta$ at which
individuals are infected, and the rate $\gamma$ at which they
recover. The model generates stochastic time-series of the number of
infected and recovered people in a population and the design problem
is to determine the times at which to measure these populations in
order to best learn the parameters (see \citet{Kleinegesse2021b} for a
detailed description of the setup). The top-left plot in Figure
\ref{fig:BOED} shows the maximisation of the lower bound in
\eqref{eq:JSD-lower-bound} and the bottom-left plot shows the
corresponding trajectories of the optimal designs. The sub-figure on
the right shows the prior and estimated posterior for data acquired at
the optimal design, i.e.\ the optimal measurement points.

If the simulator-based models are implemented such that automatic
differentiation is supported, gradient-based optimisation is
possible, which enables experimental design with higher dimensional
design vectors $\d$. Furthermore, rather than optimising with respect
to $\d$ directly, it is also possible to learn a policy which outputs
the designs $\d$, which enables adaptive sequential design for
simulator-based models \citep{Ivanova2021}.

\section{Conclusions}
The likelihood function is a main workhorse for statistical inference
and experimental design. However, it is computationally intractable
for several important classes of statistical models, including
energy-based models and simulator-based models. This makes standard
likelihood-based learning and experimental design impossible for those
models. Contrastive learning offers an intuitive and computationally
feasible alternative to likelihood-based learning. We have seen that
contrastive learning is closely related to classification, logistic
regression, and ratio estimation. By exploiting properties of the
logistic loss, we have shown how contrastive learning can be used to
solve a range of difficult statistical problems, namely (1) parameter
estimation for energy-based models, (2) Bayesian inference for
simulator-based models, and (3) Bayesian experimental design for
simulator-based models. Whilst we focused on the logistic loss, we
have pointed out that other loss functions can be used as well. The
relative benefits and possible optimality properties of the different
loss functions, as well as the reference data that is needed for
contrastive learning, remain important open research questions.

\begin{acknowledgements}
SK and BR were supported in part by the EPSRC Centre for Doctoral
Training in Data Science, funded by the UK Engineering and Physical
Sciences Research Council (grant EP/L016427/1) and the University of
Edinburgh.
\end{acknowledgements}

\section*{Conflict of interest}
The authors declare that they have no conflict of interest.

\bibliographystyle{spbasic}      % basic style, author-year citations
\bibliography{refs}   % name your BibTeX data base

\end{document}